\documentclass{article}

\usepackage{spconf,amsmath,graphicx}

\usepackage{caption}

\usepackage{float}

\usepackage{tabularx}
\usepackage{enumitem}
\setlist{nosep, leftmargin=14pt}

\usepackage{booktabs}

\usepackage{multirow}

\usepackage{makecell}

\usepackage{adjustbox}
\usepackage{siunitx}

\title{Enhancing Contrastive Learning for Retinal Imaging via Adjusted Augmentation Scales}
\name{Zijie Cheng$^{\star,1}$ \qquad Boxuan Li$^{1}$ \qquad André Altmann$^{1}$ \qquad Pearse A Keane $^{2}$ \qquad Yukun Zhou$^{\star,2}$}

\address{
    $^{1}$ UCL Department of Medical Physics \& Biomedical Engineering, United Kingdom \\
    $^{2}$ UCL Institute of Ophthalmology, United Kingdom
    \\
    $\star$ Corresponding authors. Email: \{rmapzch, yukun.zhou.19\}@ucl.ac.uk
}

\begin{document}
%
\maketitle
\begin{abstract}
Contrastive learning, a prominent approach within self-supervised learning, has demonstrated significant effectiveness in developing generalizable models for various applications involving natural images. However, recent research indicates that these successes do not necessarily extend to the medical imaging domain. In this paper, we investigate the reasons for this suboptimal performance and hypothesize that the dense distribution of medical images poses challenges to the pretext tasks in contrastive learning, particularly in constructing positive and negative pairs. We explore model performance under different augmentation strategies and compare the results to those achieved with strong augmentations. Our study includes six publicly available datasets covering multiple clinically relevant tasks. We further assess the model’s generalizability through external evaluations. The model pre-trained with weak augmentation outperform those with strong augmentation, improving AUROC from 0.838 to 0.848 and AUPR from 0.523 to 0.597 on MESSIDOR-2, and showing similar enhancements across other datasets. Our findings suggest that optimizing the scale of augmentation is critical for enhancing the efficacy of contrastive learning in medical imaging.
\end{abstract}
\section{Introduction}
\label{sec:intro}
\begin{figure*}[htbp]
    \centering
    \includegraphics[width=0.85\textwidth]{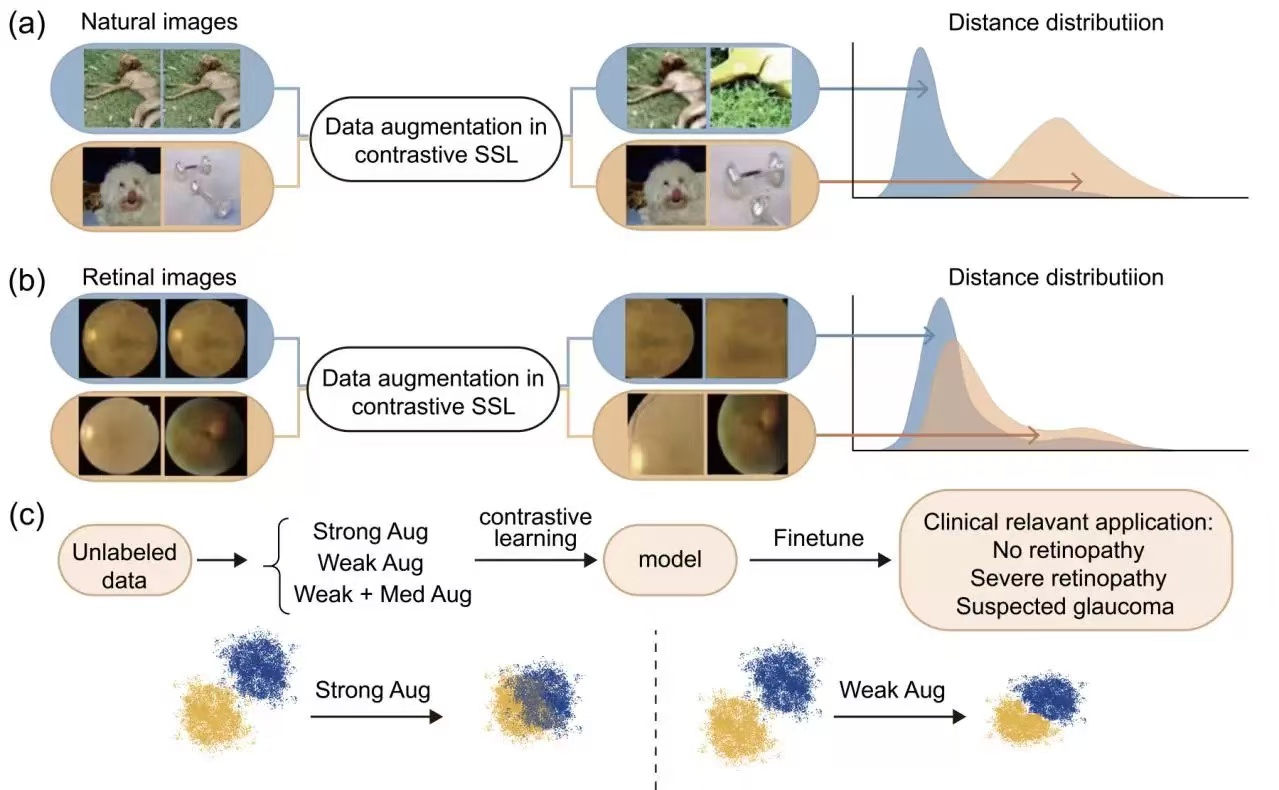}
    \caption{Figures (a) and (b) illustrate the distribution of distances between positive pairs and negative pairs in both natural and medical image domains. Figure (c) presents the project pipeline: unlabeled data is used to pre-train contrastive learning models while investigating various augmentation strategies. The blue dots and yellow dots indicate augmented images from different original images. The goal of this approach is to enhance feature clustering and improve the accuracy of retinal disease diagnosis.}
    \label{fig:project_overview}
\end{figure*}

Contrastive learning is a machine learning paradigm that trains models to distinguish between similar and dissimilar data points without relying on explicit labels. Despite being pre-trained only on unlabeled data, contrastive learning drives competitive pre-trained models compared to those pre-trained with supervised learning-based methods. In the natural image domain, it has demonstrated promising results in diverse tasks such as object detection \cite{Xie_2021_ICCV}, image classification \cite{Zeng_Xie_2021}, and video analysis \cite{Singh_2021_CVPR}. Compared to generative learning, contrastive learning has shown greater effectiveness in various applications involving natural images \cite{caron_emerging_2021,oquab2024dinov2learningrobustvisual}. However, whether this observation extends to medical images remains underexplored.

Recent research has begun comparing contrastive learning and generative learning in medical AI. For instance, RETFound \cite{zhou_foundation_2023}, a foundation model for retinal images, employed a generative learning strategy using the Masked Autoencoder \cite{he_masked_2022} for model development and demonstrated superior performance compared to contrastive learning methods in retinal disease classification. Understanding the reasons behind this inconsistency and developing a simple yet efficient solution to improve contrastive learning for medical imaging is crucial.

The suboptimal performance of contrastive learning in medical imaging is likely due to inherent differences between the distributions of natural and medical images \cite{WEN2021103145}. Natural images are colorful with varying pixel intensities, while medical images are usually grayscale and structurally similar, especially within the same organ or tissue type \cite{joy2004mr, legras2018distribution}. This characteristic results in a denser distribution of medical images within the latent space compared to natural images \cite{zhou2021review}. We hypothesize that such a dense distribution degrades performance when applying contrastive learning methods to medical images. As shown in Figures \ref{fig:project_overview}(a) and \ref{fig:project_overview}(b), natural images under strong augmentations in contrastive learning are sparsely distributed in latent space, while different medical images tend to overlap heavily. Since the pretext of contrastive learning is to differentiate between positive pairs (augmented views of the same image) and negative pairs (augmented views of different images), the severe overlap in medical images makes the pretext task highly challenging, making it difficult for the model to converge effectively. 

In this work, we propose a simple yet effective solution to enhance contrastive learning performance by reducing augmentation scales. The project pipeline is illustrated in Figure \ref{fig:project_overview}(c). We employ Distillation with No Labels (DINO) \cite{caron_emerging_2021}, and validated our solution on glaucoma and diabetic retinopathy diagnosis, using both internal and external evaluations. Our approach not only enhances feature clustering but also demonstrates improved diagnostic accuracy compared to models using stronger augmentation strategies.

\section{Methods}
\label{sec:format}

\subsection{Problem Definition}
For contrastive learning, given a set of unlabeled retinal images $\mathcal{D} = \left\{ x_i \right\}_{i=1}^{N}$, we create positive pairs $\mathcal{P}^+$ by randomly selecting an image $x_i \in \mathcal{D}$ and apply twice augmentation $\Phi_{t, s}$ respectively to get augmented data $x_i^{1}$ and $x_i^{2}$, where $t$ indicates the augmentation type and $s$ the scale range. While for negative pairs, we sample two images $x_i$ and $x_j \in \mathcal{D}$ (with $i \neq j$) and apply the augmentation to each image, forming the negative pair $\mathcal{P}^- = (x_i^{1}, x_j^{2})$. The distance between positive pairs and negative pairs can be measured by $Dis(\cdot)$:

\begin{equation}
\operatorname{Dis}(\mathcal{P}^+) = | x_i^{1} - x_i^{2} |,
\label{eq:positive_euclidean_distance}
\end{equation}

\begin{equation}
\operatorname{Dis}(\mathcal{P}^-) = | x_i^{1} - x_j^{2} |.
\label{eq:negative_euclidean_distance}
\end{equation}

The general training objective of contrastive learning is to train the model $f$ to maximize the distance between  negative pairs and to minimize that for positive pairs, 

\begin{equation}
f = \operatorname*{argmax} \left( \text{Dis}(\mathcal{P}^-) - \text{Dis}(\mathcal{P}^+) \right).
\label{eq:model optimization}
\end{equation}

When $\text{Dis}(\mathcal{P}^+)$ approximates $\text{Dis}(\mathcal{P}^-)$, it is challenging to train the model $f$ to converge well. This issue is prominent in medical imaging due to less variation compared to natural images. For example, retinal images depict the anatomical tissue of retina, often showing similar structure and orientation \cite{PATTON200699}, as shown in Figure \ref{fig:project_overview}(b). With strong augmentations $\Phi_{strong}$ (e.g., cropping the images into small patches), $\text{Dis}(\mathcal{P}^-)$ decrease while $\text{Dis}(\mathcal{P}^+)$ increases, which brings further challenges in achieving objects of equation \ref{eq:model optimization} and may result in suboptimal model pre-training with contrastive learning, showing the poor performance in classifying the positive and negative pairs. 

Such suboptimal model performance extends to downstream applications, where models are fine-tuned with labeled data \( \mathcal{D}_l = \left\{ x_i, y_i \right\}_{i=1}^{L} \) for diverse tasks like disease diagnosis, where \( x \) represents the data and \( y \) indicates the label. To improve the model's capability in clinically meaningful applications, we aim to optimize \( \arg\max_{x_i, y_i \in \mathcal{D}_l} \, T\left( E(x_i), y_i \right) \), where \( T(\cdot) \) is the score function and \( E(\cdot) \) is the encoder of the model \( f \). 

Our strategy involves enhancing the contrastive learning performance by specifically decreasing \( \text{Dis}(\mathcal{P}^+) \) while increasing \( \text{Dis}(\mathcal{P}^-) \).

\subsection{Scattering Data Distribution with Weak Augmentations}

To achieve such a goal for retinal images, a straightforward strategy is to scale down the augmentation. An extreme case is to remove the augmentation so that $\text{Dis}(\mathcal{P}^+)$ achieves 0 and $\text{Dis}(\mathcal{P}^-)$ stays as a high value. However, pre-training without any augmentation hardly trains the model to learn generalizable and diverse features. Hence, we propose to proportionally scale down the augmentation, termed as $\Phi_{weak}$, to ease the challenge of training model $f$ to converge while also allowing the model to learn generalizable features. Additionally, we also investigate the effects of several augmentations that mimic the retinal image artefacts, including random bias field, and Gaussian blur. We combine it with $\Phi_{weak}$ to form $\Phi_{weak+med}$. 

\section{EXPERIMENT}
\label{sec:pagestyle}

\subsection{Data}
We evaluate the efficacy of different augmentation strategies using clinically meaningful tasks, including diabetic retinopathy (DR) diagnosis, glaucoma detection, and multi-class retinal disease classification. 

\begin{table}[t]
    \scriptsize 
    \centering
    \caption{Data summary for the datasets used for disease diagnosis. Each dataset is split into training, validation, and testing sets.}
    \label{tab:datasets}
    \begin{tabular}{>{\centering\arraybackslash}p{1.6cm} >{\centering\arraybackslash}p{0.9cm} >{\centering\arraybackslash}p{0.6cm} >{\centering\arraybackslash}p{0.9cm} >{\centering\arraybackslash}p{0.9cm} >{\centering\arraybackslash}p{0.9cm}}
        \toprule
        \textbf{Dataset} & \textbf{Country} & \textbf{Types} & \textbf{Training} & \textbf{Validation} & \textbf{Testing} \\
        \midrule
        \multicolumn{6}{c}{\textbf{Diabetic retinopathy}} \\
        \midrule
        MESSIDOR-2 & France & 5 & 972 & 246 & 526 \\
        IDRiD & India & 5 & 329 & 84 & 103 \\
        APTOS2019 & India & 5 & 2048 & 514 & 1100 \\
        \midrule
        \multicolumn{6}{c}{\textbf{Glaucoma}} \\
        \midrule
        PAPILA & Spain & 3 & 312 & 79 & 98 \\
        \midrule
        \multicolumn{6}{c}{\textbf{Multi-class disease}} \\
        \midrule
        JSIEC & China & 39 & 534 & 150 & 316 \\
        Retina & NR & 4 & 336 & 84 & 181 \\
        \bottomrule
    \end{tabular}
\end{table}

For DR diagnosis, we include MESSIDOR-2 \cite{decenciere2014feedback}, IDRiD \cite{IDRiD}, and APTOS2019 \cite{APTOS2019}. The labels for DR are based on the International Clinical DR Severity Scale, covering five stages from no DR to proliferative DR. For glaucoma diagnosis, we use the PAPILA dataset \cite{kovalyk_papila_2022}, which has three categorical labels: non-glaucoma, early glaucoma (suspected glaucoma), and advanced glaucoma. For multi-class disease classification tasks, we use two datasets, JSIEC \cite{cen_automatic_2021} containing 1,000 images with 39 categories of common retinal diseases and conditions, and Retina dataset \cite{Retina} with labels for normal, glaucoma, cataract, and retinal disease. Data splitting details are shown in Table \ref{tab:datasets}. The pre-training data are from Moorfields Eye Hospital \cite{zhou_foundation_2023}.

\subsection{Pre-training details} 

DINO \cite{caron_emerging_2021}, a representative and commonly used contrastive learning strategy, was used in the experiment. We first initialized the model with ImageNet weights and then pre-trained it using 1.4 million retinal images from Moorfields Eye Hospital. The data preprocessing, data quality control, model architecture, and hyperparameters (except for those related to augmentations) were standardized to ensure a fair comparison. The model was pre-trained using an NVIDIA A100 (80G). The details of $\Phi_{strong}$, $\Phi_{weak}$, and $\Phi_{weak+med}$ are listed in Table \ref{table:augmentation_}. Local crop indicates the range of cropping local patches and global crop represents that for cropping global patches. Color jittering involves random adjustments to image brightness and contrast to simulate variations in imaging conditions. Gaussian blur applies a Gaussian filter to smooth images, mimicking motion blur or out-of-focus effects. Random noise adds Gaussian noise to simulate sensor or acquisition noise. Random bias field introduces smooth, spatially varying intensity variations to mimic changes in light illumination direction.

We then compared these models by adapting them to downstream tasks, i.e., disease diagnosis. We evaluated the model performance with the Area Under the Receiver Operating Characteristic curve (AUROC) and the Area Under the Precision-Recall curve (AUPR). Each experiment is run five times with random seeds to obtain performance statistics.

\begin{table}[t]
\centering
\caption{Various settings of augmentation types and scales. Augmentations not listed are consistent with the strong augmentations. For local and global crops, the range (e.g., (0.05, 0.4)) represents the cropping scales relative to the original image. The symbol p denotes the probability of applying a particular transformation. A "\(\times\)" indicates that the transformation is not applied.}
\label{table:augmentation_}
\renewcommand{\arraystretch}{1.8} 
\setlength{\tabcolsep}{2pt}       
\fontsize{10pt}{12pt}\selectfont  
\begin{adjustbox}{width=0.48\textwidth}
\begin{tabular}{@{}l p{1.8cm} p{1.8cm} p{2.5cm} p{1.2cm} p{1.2cm} p{1.2cm}@{}}
\toprule
 & Local crop & Global crop & Color jitter & Blur & Noise & Bias field \\
\midrule
$\Phi_{\text{strong}}$ & (0.05, 0.4) & (0.4, 1.0) & \makecell[{{p{2.5cm}}}]{bright:0.4 \\ contrast:0.4} & $\times$ & $\times$ & $\times$ \\
\midrule
$\Phi_{\text{weak}}$ & (0.2, 0.5) & (0.5, 1.0) & \makecell[{{p{2.5cm}}}]{bright:0.2 \\ contrast:0.2} & $\times$ & $\times$ & $\times$ \\
\midrule
$\Phi_{\substack{\text{weak} \\ + \text{med}}}$ & (0.2, 0.5) & (0.5, 1.0) & \makecell[{{p{2.5cm}}}]{bright:0.2 \\ contrast:0.2} & \makecell[{{p{1.2cm}}}]{std:0.1 \\ p:0.5} & \makecell[{{p{1.2cm}}}]{std:0.1 \\ :0.5} & \makecell[{{p{1.2cm}}}]{scale:0.1 \\ :0.5} \\
\bottomrule
\end{tabular}
\end{adjustbox}
\end{table}

\begin{table*}[t]
\centering
\caption{This table presents the external evaluation results on diabetic retinopathy (DR) datasets based on AUROC. For each dataset pair, the highest mean value among the different augmentation strategies is highlighted in bold.}
\label{table:external_eval}
\begin{adjustbox}{width=0.8\textwidth}
\begin{tabular}{llcccccc}
\hline
\textbf{Fine-tune data} & & \multicolumn{2}{c}{APTOS2019} & \multicolumn{2}{c}{IDRiD} & \multicolumn{2}{c}{MESSIDOR-2} \\
\hline
\textbf{Test data} & & IDRiD & MESSIDOR-2 & APTOS2019 & MESSIDOR-2 & APTOS2019 & IDRiD \\
\hline
$\Phi_{strong}$ & & .784 $\pm$ .003 & \textbf{.767 $\pm$ .006} & .745 $\pm$ .016 & .742 $\pm$ .031 & \textbf{.806 $\pm$ .023} & .740 $\pm$ .034 \\
$\Phi_{weak}$ & & \textbf{.790 $\pm$ .019} & .760 $\pm$ .007 & \textbf{.749 $\pm$ .033} & \textbf{.760 $\pm$ .027} & .798 $\pm$ .038 & \textbf{.744 $\pm$ .019} \\
$\Phi_{weak + med}$ & & .751 $\pm$ .014 & .691 $\pm$ .002 & .733 $\pm$ .055 & .720 $\pm$ .020 & .706 $\pm$ .077 & .736 $\pm$ .034 \\
\hline
\end{tabular}
\end{adjustbox}
\end{table*}

\subsection{Experiment Result}
\begin{figure}[t]
    \centering
    \includegraphics[width=0.48\textwidth]{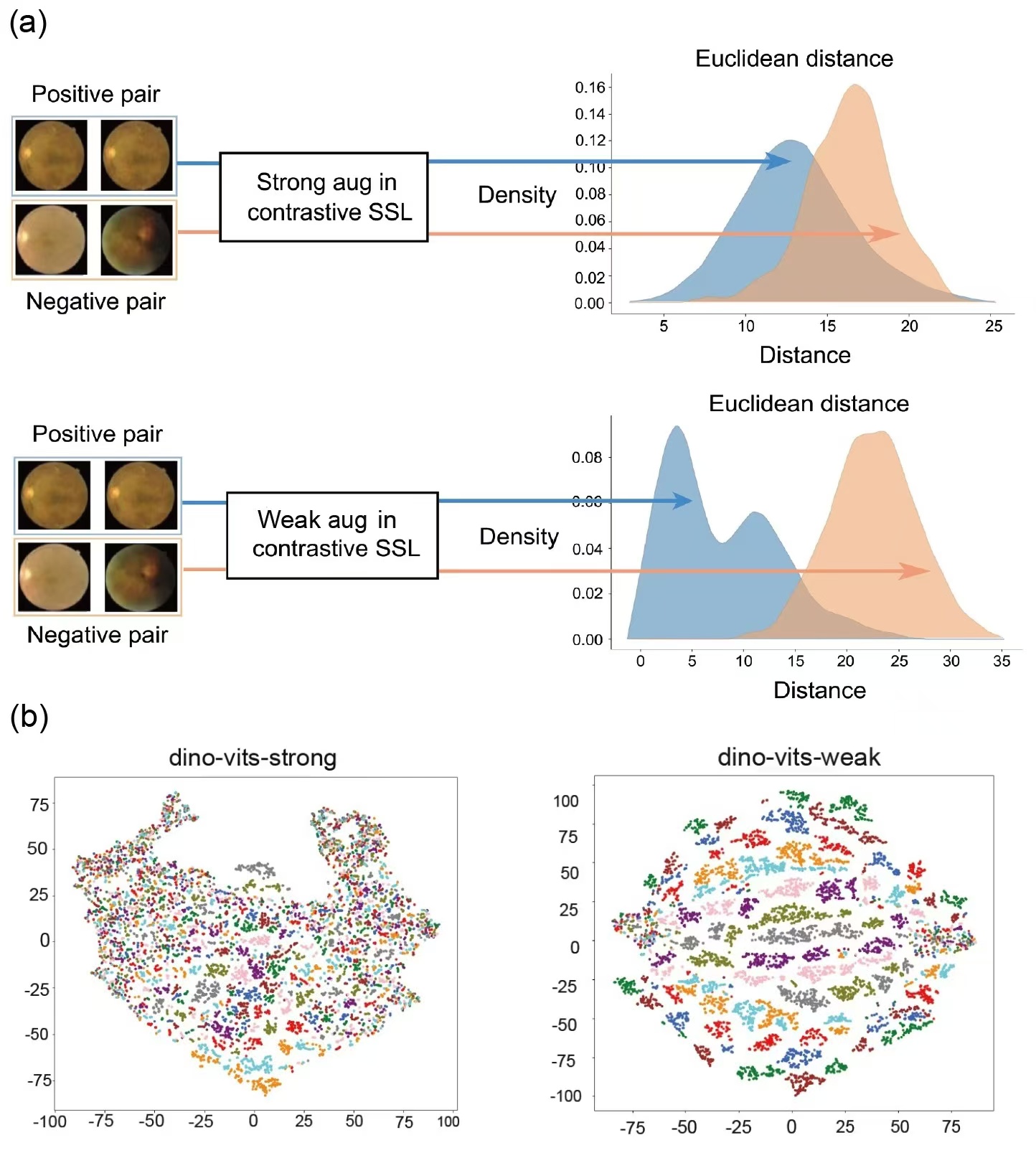}
    \caption{We extract features using the DINO teacher model (encoder), pre-trained separately with strong and weak augmentations. First, we calculate the Euclidean distances between positive and negative pairs and compare their distance distributions in Figure (a). We also use a t-SNE map to visualize the feature clustering in Figure (b), where different colors represent augmented views from different images.}
    \label{fig:dino_t-sne_and_data_distribution}
\end{figure}

We first observed the clustering performance, i.e., how positive and negative pairs distribute, before and after reducing the augmentation scale. The results are shown in Figure \ref{fig:dino_t-sne_and_data_distribution}. The model pre-trained with $\Phi_{weak}$ better separated these pairs. We also repeatedly augmented each image multiple times to create image groups, where positive pairs consist of images within the same group, and negative pairs are images from different groups. We found that positive pairs cluster more closely under weak augmentation in the t-SNE map \cite{maaten2008visualizing}.
\begin{table}[t]
\centering
\caption{justificson on disease diagnosis with internal evaluation. Each column indicates the model pretrained with varied data augmentation strategies. The highest value in each row is highlighted in bold.}
\label{table:internal_eval}
\begin{adjustbox}{width=0.48\textwidth}
\begin{tabular}{lccc}
\hline
\phantom{Measurement}         & $\Phi_{strong}$                 & $\Phi_{weak}$                   & $\Phi_{weak + med}$             \\
\hline
\multicolumn{4}{l}{\textbf{MESSIDOR-2}} \\
\hline
AUROC               & .838 \,(.834, .842) & \textbf{.848 \,(.844, .852)} & .823 \,(.814, .832) \\
AUPR                & .523 \,(.513, .533) & \textbf{.597 \,(.583, .610)} & .523 \,(.487, .558) \\
\hline
\multicolumn{4}{l}{\textbf{APTOS2019}} \\
\hline
AUROC               & .933 \,(.932, .933) & .933 \,(.931, .935) & .924 \,(.924, .925) \\
AUPR                & \textbf{.667 \,(.664, .670)} & .662 \,(.655, .669) & .637 \,(.635, .640) \\
\hline
\multicolumn{4}{l}{\textbf{IDRiD}} \\
\hline
AUROC               & .747 \,(.731, .763) & \textbf{.790 \,(.778, .802)} & .726 \,(.717, .734) \\
AUPR                & .461 \,(.439, .482) & \textbf{.498 \,(.481, .514)} & .432 \,(.413, .452) \\
\hline
\multicolumn{4}{l}{\textbf{PAPILA}} \\
\hline
AUROC               & .791 \,(.778, .803) & \textbf{.816 \,(.799, .834)} & .792 \,(.782, .803) \\
AUPR                & .637 \,(.628, .646) & \textbf{.671 \,(.646, .696)} & .628 \,(.615, .641) \\
\hline
\multicolumn{4}{l}{\textbf{JSIEC}} \\
\hline
AUROC               & .960 \,(.957, .963) & \textbf{.977 \,(.974, .979)} & .968 \,(.966, .970) \\
AUPR                & .651 \,(.631, .670) & \textbf{.760 \,(.746, .773)} & .707 \,(.690, .725) \\
\hline
\multicolumn{4}{l}{\textbf{Retina}} \\
\hline
AUROC               & .781 \,(.774, .789) & .807 \,(.799, .815) & \textbf{.814 \,(.804, .823)} \\
AUPR                & .594 \,(.580, .608) & \textbf{.632 \,(.615, .648)} & .626 \,(.612, .639) \\
\hline
\end{tabular}
\end{adjustbox}
\end{table}

In internal evaluation, as shown in Table \ref{table:internal_eval}, DINO with $\Phi_{weak}$ outperforms other augmentation strategies in most retinal disease classification tasks. On MESSIDOR-2, IDRiD, PAPILA, and JSIEC, the model with $\Phi_{weak}$ shows consistently better performance in both AUROC and AUPR metrics compared to with $\Phi_{strong}$. Notably, on JSIEC, the model pre-trained with $\Phi_{weak}$ achieves a 10\% higher AUPR than with $\Phi_{strong}$. However, when medical augmentation $\Phi_{med}$ is introduced, the model’s performance decreases in most cases. On MESSIDOR-2 and IDRiD, $\Phi_{weak+med}$ reduces the model’s performance by 2.5\% and 6.4\%, respectively, even making it lower than with $\Phi_{strong}$. 

For external evaluation, as shown in Table \ref{table:external_eval}, $\Phi_{weak}$ also achieved a marginal improvement in terms of model generalizability compared to $\Phi_{strong}$. When the model trained on IDRiD was externally evaluated on APTOS2019 and MESSIDOR-2, the model pre-trained with $\Phi_{weak}$ outperforms $\Phi_{strong}$ by 0.4\% and 1.8\%, respectively.

\section{Conclusion}

In this study, we aimed to improve the contrastive learning performance in the medical image domain. We proposed a hypothesis that the dense distribution of medical images might cause the suboptimal performance of contrastive learning, and validated in our experiments and validation. Our findings suggest that simply reducing augmentation scales to an appropriate level can improve the clustering performance and therefore enhance model performance. Additionally, when incorporating medical-specific augmentation $\Phi_{med}$ to $\Phi_{weak}$, the collective augmentation again decreases $\text{Dis}(\mathcal{P}^-)$, while increase $\text{Dis}(\mathcal{P}^+)$, generating adverse effects on model performance. These offer key guidance into the model pre-training with contrastive learning for medical images.

Although bringing insights, we acknowledge several limitations in this work that should be studied in future work. First, we only validated our hypothesis and solution on DINO; more contrastive learning strategies, such as DINOv2 \cite{oquab2024dinov2learningrobustvisual}, could be investigated. Second, some quantitative metrics describing the clustering performance have not been investigated, which will be proposed in future work to guide the augmentation scaling. Finally, some techniques like tailored loss functions adjusting the weights on positive and negative pairs will be studied. This work pioneered the optimization of contrastive learning in the medical domain and encouraged the tailored model training settings for medical images.

\section{Compliance with Ethical Standards}
Only de-identified retrospective data were used in this research, without the active involvement of patients. All datasets used in the downstream applications are publicly available.

\section{Acknowledgment}

No funding was received for conducting this study. The authors have no relevant financial or non-financial interests to disclose.

\bibliographystyle{IEEEbib}
\bibliography{strings,main}

\end{document}